\documentclass[final]{cvpr}

\usepackage{times}
\usepackage{epsfig}
\usepackage{graphicx}
\usepackage{amsmath}
\usepackage{amssymb}
\usepackage{cite}
\usepackage{bbding}
\usepackage{multirow}
\usepackage{subfigure}
\usepackage{booktabs}
\usepackage{multirow}
\usepackage{booktabs}
\usepackage{bbding}
\usepackage{pifont}
\usepackage{wasysym}
\usepackage{amssymb}
\usepackage{arydshln}

\usepackage[pagebackref=true,breaklinks=true,colorlinks,bookmarks=false]{hyperref}

\def\cvprPaperID{69} 
\def\confYear{CVPR 2021}

\begin{document}

\title{Toward Interactive Modulation for Photo-Realistic Image Restoration}

\author{Haoming Cai$^{1 \ *}$ \qquad\quad Jingwen He$^{1}$ \thanks{Denotes equal contribution} \qquad\quad Yu Qiao$^{1,2}$ \qquad\quad Chao Dong$^{1,3}$ \thanks{Denotes corresponding author (e-mail: chao.dong@siat.ac.cn)} \\
$^{1}$Key Laboratory of Human-Machine Intelligence-Synergy Systems, \\ Shenzhen Institutes of Advanced Technology, Chinese Academy of Sciences\\
$^{2}$Shanghai AI Lab, Shanghai, China\\
$^{3}$SIAT Branch, Shenzhen Institute of Artificial Intelligence and Robotics for Society\\
{\tt\small \{hm.cai, jw.he, yu.qiao, chao.dong\}@siat.ac.cn}
}

\maketitle

\pagestyle{empty}  
\thispagestyle{empty} 

\begin{abstract}
Modulating image restoration level aims to generate a restored image by altering a factor that represents the restoration strength.
Previous works mainly focused on optimizing the mean squared reconstruction error, which brings high reconstruction accuracy but lacks finer texture details.
This paper presents a Controllable Unet Generative Adversarial Network (CUGAN) to generate high-frequency textures in the modulation tasks.
CUGAN consists of two modules - base networks and condition networks. 
The base networks comprise a generator and a discriminator. 
In the generator, we realize the interactive control of restoration levels by tuning the weights of different features from different scales in the Unet architecture. 
Moreover, we adaptively modulate the intermediate features in the discriminator according to the severity of degradations. 
The condition networks accept the condition vector (encoded degradation information) as input, then generate modulation parameters for both the generator and the discriminator. 
During testing, users can control the output effects by tweaking the condition vector.
We also provide a smooth transition between GAN and MSE effects by a simple transition method.  
Extensive experiments demonstrate that the proposed CUGAN achieves excellent performance on image restoration modulation tasks.
\end{abstract}

\section{Introduction}

\begin{figure}[t]
	\begin{center}
		\includegraphics[width=1.0\linewidth]{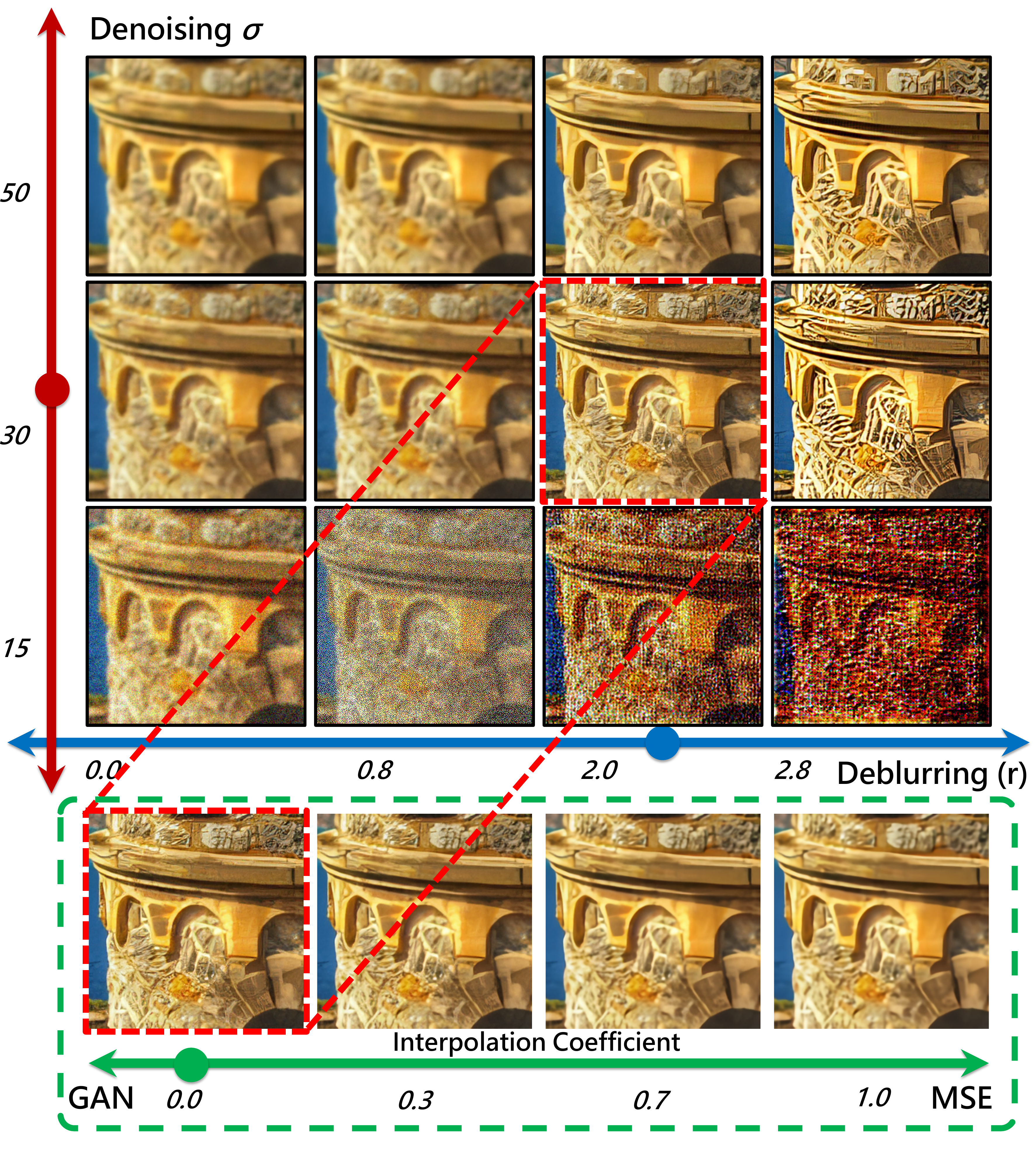}
	\end{center}
	\vspace{-1em}
	\caption{The input image is with blur $r2$ (kernel width=2) and $\sigma 30$. Through the modulation on deblurring and denoising, users could obtain a well restored image with rich texture details. Besides, users are allowed to modulate on the green sliding bar to achieve smooth transition between GAN and MSE effects.}
	\label{figure1}
	\vspace{-1.0em}
\end{figure}

\begin{figure}[t]
	\begin{center}
		\includegraphics[width=1.0\linewidth]{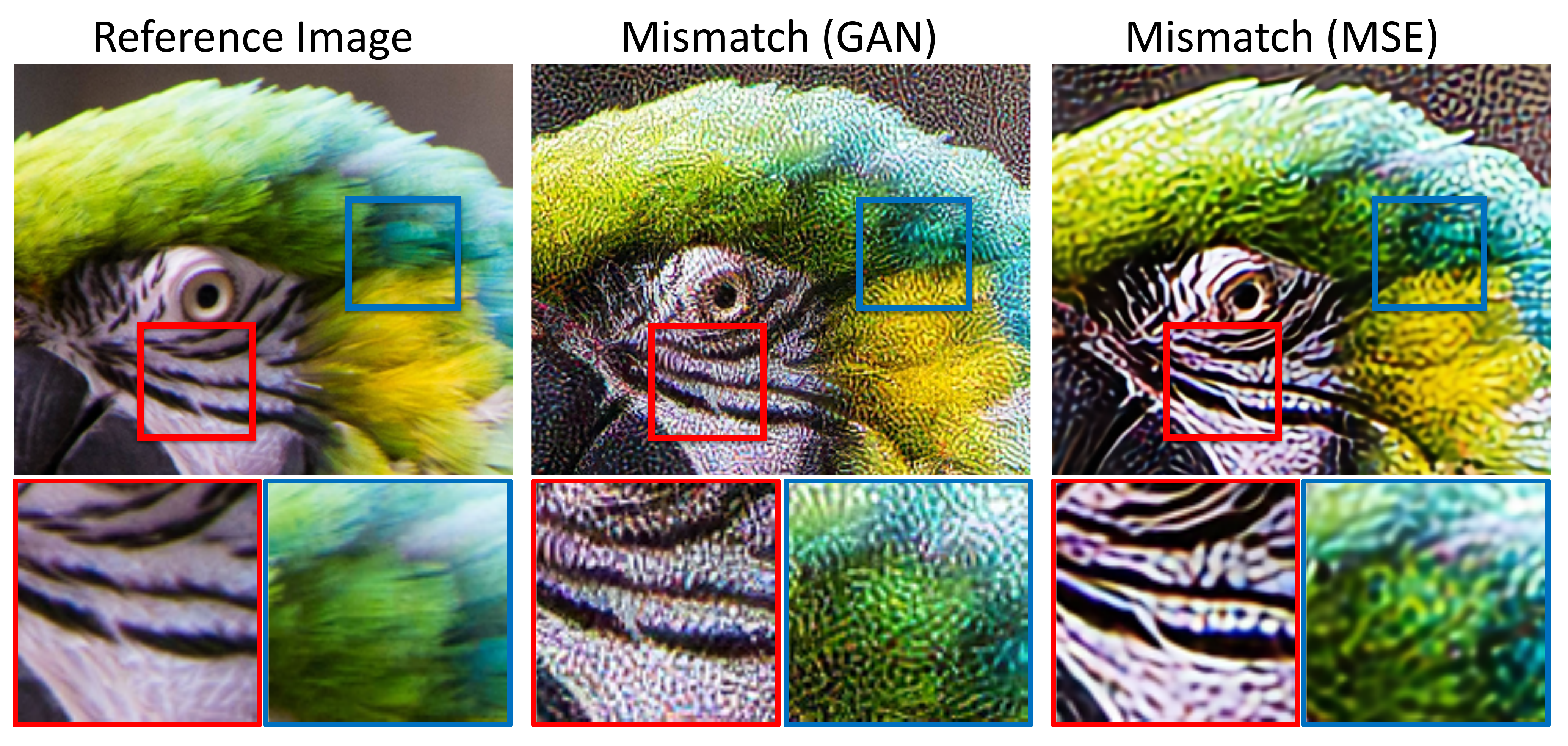}
	\end{center}
	\caption{Artifacts incurred by applying GAN-based and PSNR-oriented model with mismatched restoration level. Models are trained on blur $r2$+noise
$\sigma30$, while the input image is with blur $r1$+noise $\sigma 50$.}
	\label{mismatch}
	\vspace{-1em}
\end{figure}

Deep learning methods have achieved great success in many low-level vision tasks, such as image denoising, deblurring, and super-resolution. 
Various network architectures and training strategies have been continuously improving the reconstruction quality (e.g., PSNR). 
Later on, to pursue visually pleasing results, the generative adversarial network (GAN) \cite{gan} is introduced to encourage the network to produce natural looking images. 
For most image restoration tasks, a deep model learns a deterministic mapping and outputs a fixed result for a pre-determined degradation type/level.
For instance, many image restoration deep models are trained for a single level (e.g., Gaussian noise $\sigma$50). 
However, this deterministic mapping is not flexible, as users cannot continuously modify the restoration effect or strength based on personal preference.
Moreover, when the degradation model mismatches the degraded input, the output image will contain severe artifacts, especially for GAN-based models, as shown in Figure \ref{mismatch}. 
Therefore, developing modulation models that can flexibly handle a range of degradations by sliding bars is essential and practical.

In recent years, several modulation methods \cite{AdaFM,DNI,cfsnet,decouple} have been proposed to adapt existing deep models to other objectives. 
Specifically, they can generate continuous restoration results between the pre-defined start level and end level (e.g, denoising $\sigma15\rightarrow\sigma50$). 
Furthermore, CResMD ~\cite{cresmd} proposes a multi-dimension modulation framework that allows jointly modulation for different degradations.
However, the above modulation methods are all PSNR-oriented, which will produce over-smoothed results without sufficient high-frequency details.
To obtain modulation outputs with photo-realistic effect, this work focuses on interactive modulation for GAN-based image restoration.


The main challenges behind PSNR-oriented and GAN-based modulation are different.
For PSNR-oriented modulation methods, such as CResMD, severe and mild degradations have different magnitude orders on the MSE loss, which will lead to the "unbalanced learning" problem. 
This phenomenon encourages models to focus on restoring images with severe degradations while ignoring the mild ones.
On the contrary, in GAN-based modulation, the generator may ignore those severe degradations if we directly apply a vanilla GAN.
Specifically, for a vanilla discriminator, an image restored from severe degradations will look like the fake one compared with that from mild ones.
This incorrect judgment will lead to the vanishing generator gradient on those severe degradations.
Therefore, we need specific discriminator to clarify images restored from various degradations.
This discriminator could adjust its judgment criterion based on the degradation of the input image.

In this paper, we introduce a novel GAN-based modulation framework for photo-realistic image restoration.
The proposed controllable Unet generative adversarial network (CUGAN) adopts the Unet architecture.
We realize the interactive control of the reconstruction result by tuning the weights of different features from different scales or within one scale in the Unet framework.
To tackle vanishing generator gradient on severe degradations, we apply global feature modulation (GFM) \cite{CSRNet} in the discriminator.
Based on the severity of degradations, the global feature modulation adaptively modulates intermediate features of the discriminator.
The above modulation is achieved by the condition networks embedded in the CUGAN.  
The condition networks accept a condition vector that encodes the degradation information for each input image.
Then, the condition networks generate the tuning weights for the generator and the parameters of global feature modulation (GFM) for the discriminator. 
During testing, users can control the restoration strength by tweaking the condition vector.
We also provide the trade-off method between GAN and MSE effect, allowing users to adjust the output effect in another dimension.
Different from DNI \cite{DNI}, our interpolation works on both the base network $G$ and the condition network $C_{G}$. 
Figure \ref{figure1} shows a qualitative example for the whole modulation process.

To verify the effectiveness of the proposed method, we conduct extensive experiments on modulation for image restoration with multiple degradations.
Experimental results show that the proposed CUGAN could achieve modulation with high visual quality, high reconstruction accuracy, or a compromise between them. 
We also demonstrate its effectiveness on image restoration in real-world scenarios.
In this work, our main contributions can be summarized as follows:
\begin{itemize}
\item[$\bullet$]
We propose a novel GAN-based modulation framework, named CUGAN, to obtain photo-realistic restoration results by interactive modulation.
\item[$\bullet$]
We propose a discriminator equipped with global feature modulation to ease the vanishing generator gradient on severe degradations.
\item[$\bullet$]
We test the restoration performance on selected datasets and real-world images. Extensive experiments show the effectiveness of CUGAN in modulated image restoration.
\end{itemize}

\section{Related Work}
\textbf{Perceptual Image Restoration.}
With the rapid development of deep learning in recent years, deep learning techniques have been widely explored to tackle image restoration problems, such as image super-resolution, denoising, deblurring, and compression artifacts reduction \cite{dncnn,ffdnet,cbdnet,arcnn,zhang2019deep,IKC}. 
Focusing on optimizing PSNR, the above methods tend to generate blurry images. 
In contrast, perceptual image restoration targets to obtain better perceptual results with more texture details by appling GAN\cite{gan}. 
Ledig et al. \cite{srgan} propose SRGAN that could generate photo-realistic images in SR task. 
In the PIRM2018-SR Challenge \cite{PIRM}, ESRGAN \cite{esrgan} achieves state-of-the-art performance by improving the network architecture and loss functions. 
Benefiting from a learnable ranker, RankSRGAN \cite{ranksrgan} could generate visually pleasant images that favor different perceptual metrics. 

\textbf{Explorative Image Restoration.}
Despite that deep-learning-based methods have achieved high qualitative performance, most of them deal with image restoration problem by learning a deterministic mapping.
To allow users adjusting the restoration effects, Bahat et al.\cite{bahat2020explorable} propose an editing module that could iteractively influence the texture or brightness.
With normalizing flows \cite{dinh2016density,kingma2018glow}, SR-Flow \cite{lugmayr2020srflow} takes a step forward to model the conditional distribution of all possible SR reconstructions given an LR facial input.
Furthermore, DeepSEE \cite{buhler2020deepsee} also adopts normalizing flow to leverage semantic maps for explorative facial super-resolution.

\textbf{Modulation for Image Restoration.}
Although existing methods could allow user to adjust restored effect, most image restoration deep networks are trained on one specific degradation level.
Therefore, it is tiring to train $N$ various models for $N$ various degradations.
DNI \cite{DNI} and AdaFM \cite{AdaFM} find the high similarity on kernels between models trained on various levels.
Based on this observation, DNI directly interpolate parameters between two related networks to attain a smooth control of diverse imagery effects.
While AdaFM adopts a more efficient way: utilize depth-wise convolution layers to modulate the intermediate features instead of changing all convolution filters. 
Different from above two interpolation-based methods, the work in CFSnet \cite{cfsnet} adaptively learns the interpolation coefficients and uses them to couple intermediate features from the main branch and tuning branch.

However, In real-world scenarios, images contain multiple degradations, such as blur and noise. Since different types of degradations are coherently related, modulation for multiple degradations should be conducted jointly, not independently. 
To address this, CResMD \cite{cresmd} regards the modulation as a conditional image restoration problem, and proposed a framework that accepts both corrupted images and their degradation information as input.
Thus, image restoration is conditioned on the restoration/degradation information.
Modulation for image restoration is essential in practical usages not merely because that it enables a unified framework to handle multiple degradations with arbitrary levels. 
More importantly, it provides a flexible interaction on the restoration strength.




\begin{figure*}[htbp]
\vspace{-2.5em}
\centering  
\includegraphics[width=1.0\linewidth]
{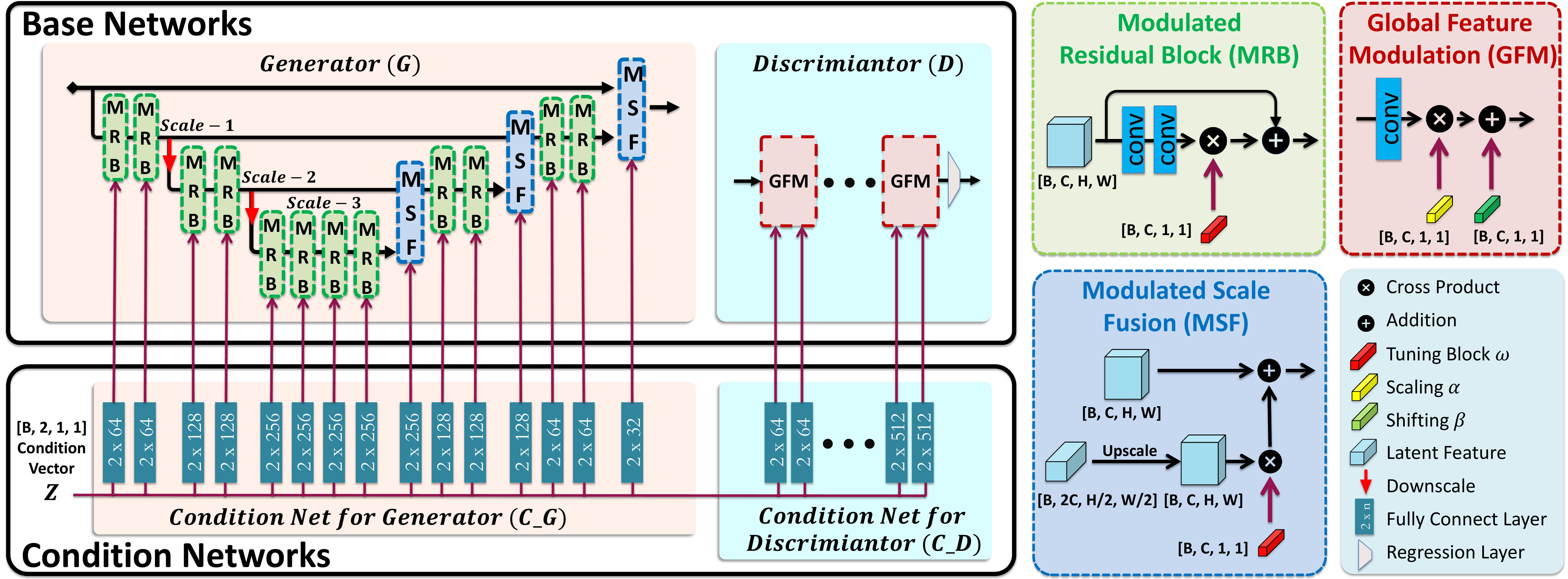}  
\caption{The framework of CUGAN.
CUGAN consists of 2 base networks as well as 2 condition networks. 
The 2 base networks are the generator $G$ and the discrimiantor $D$.
The 2 condition networks are condition network for generator ($C_{G}$) and condtion network for discriminator ($C_{D}$).
During training and testing process, the condition networks $C_{G}$ and $C_{D}$ accept degradation information and generate tunning weights for $G$ and $D$. 
In the modulation Unet, we provide Modulated Residual Block (MRB) and Modulated Scale Fusion (MSF) to modulate features from different scales or within one scale.
As for the modulation discriminator, we provide Global Feature Modulation (GFM) to scale and shift intermediate feature maps.
} 
\label{figure_framework}
\vspace{-1.0em}
\end{figure*}

\section{Methods}
Our goal is to design a GAN-based restoration model that takes in both the degraded image and desired restoration information as inputs and outputs visually pleasing restored image. 
The restoration information, which is equal to the degradation information, can be regarded as sliding bars for users to modulate during testing.
We achieve such a photo-realistic modulation model by the following approach. Given any corrupted image $I^{dis}_{i}$ distorted with some degradations, we want to restore it to a clear image $I^{res}_{i}$, which is close to the ground truth image $I^{gt}$. To allow image restoration for different degradations, we accept a condition vector $z$ ($z^k\in [0, 1]$). In particular, each dimension of $z$ represents a certain degradation type while its value encodes the degradation level. Therefore, our task is to find a function $F$, such that: $F(I^{dis}_{i}, z_i) \rightarrow I^{gt}$, where $i=1,2 \ldots, N$, and $N$ indicates the number of all predefined degradations. In addition, we introduce a conditional discriminator, that aims to distinguish between the restored image $I^{res}_{i}$ and the ground truth image $I^{gt}$ under condition of $z_i$. In particular, the discriminator accepts $I^{res}_{i}$, $I^{gt}$, and $z_i$ as inputs, and outputs the two probabilities for restored result and ground truth image, respectively.

\subsection{Base Networks}
The base networks consist of two parts: generator $G$ and discriminator $D$. 
The generator $G$ accepts the input image and outputs the restored result, while $D$ aims to discriminate the restored result from the ground truth.

\noindent{\textbf{Generator.}} 
The architecture of the generator $G$ is shown in Figure \ref{figure_framework}. 
Basically, it is a Unet framework incorporated with Residual Blocks. 
$G$ consists of three scales, namely $scale_1$, $scale_2$, and $scale_3$. 
This multi-scale architecture could help achieve better results for GAN-based modulation compared with a plain network (Please see the experimental results in Ablation Study).
For each scale, there are two residual blocks in the left and the other two in the right. 
Each residual block contains two $3\times3$ convolution layers, and a ReLU activation function between them. 
In $scale_1$, there are two $3\times3$ convolution layers at two ends.
As for $scale_2$ and $scale_3$, we begin by using $2\times2$ strided convolution to downscale the features and end up with $2\times2$ transposed convolution for upscaling. 
The number of channels for each layer from $scale_1$, $scale_2$, $scale_3$ are set to 64, 128, and 256, respectively. 
Besides, each scale has a short connection that could skip its successive scale.
Note that we also add a global connection to ease the difficulty for restoration on mild degradations.

\noindent{\textbf{Discriminator.}}
The discriminator $D$ is to discriminate ground truth images from restored images. 
It accepts 64$\times$64 image patches, and outputs the predictions.
The architecture of $D$ is shown in Figure \ref{figure_framework}. 
We follow the work in \cite{srgan}, and design a discriminator that progressively down-sample the feature maps to a feature vector. 
Specifically, we use several strided convolution layers to half the sizes of feature maps each time. 
On the other hand, we increase the channels of feature maps by using convolution layers with doubled number of filters. 
There are 10 convolution layers in total. 
LeakyReLU \cite{lrelu} activation is adopted between convolution layers. 
After we obtaining the feature maps with 512 channels, we use global average pooling and two fully-connected layers (regression) to output the final probability.

\subsection{Condition networks}
The condition networks mainly contain two parts, one for the generator and the other for the discriminator, namely as $C_{G}$ and $C_{D}$, respectively.
Each condition network accepts a condition vector that encodes the degradation information for the input image, and outputs the parameters of modulation operations that will be used to modulate the intermediate feature in the base networks $G$ and $D$.

First, the degradation information of each corrupted image $x$ should be encoded into the condition vector $z$. 
Specifically, for each degradation type, the corresponding degradation level is scaled to a value within range $[0, 1]$.
For instance, given an input image with blur level $r=1$ ($r \in [0, 4]$) and noise level $\sigma=10$ ($\sigma \in [0, 50]$), we could obtain a corresponding condition vector $z=[0.25, 0.20]$ by computing $[1/4, 10/50]$.

The architectures of condition networks are shown in Figure \ref{figure_framework}. 
Each condition network consists of several independent fully-connected layers (FC).  
To generate the tuning weights for the generator, the $i$-th fully-connected layer transforms the condition vector $z$ of the input image $x$ to the tuning weight $w_i$ for the $i$-th modulation module in the generator. 
The formulation can be written as follows:
\begin{align}
w_i = F^w_i(z),\nonumber
\end{align}
where $F^w_i(\cdot)$ denotes the function of $i$-th fully-connected layer. The dimension of $w_i$ is the same as the channel number of the corresponding feature maps to be modulated.

As for the $i$-th modulation module in the discriminator, we use two independent fully-connected layers to respectively generate the parameters of scaling and shifting operations for global feature modulation (GFM). In particular, we have:
\begin{align}
\alpha_i = F^{\alpha}_{i}(z),\nonumber
\beta_i = F^{\beta}_{i}(z),\nonumber
\end{align}
where $F^{\alpha}_{i}(\cdot)$ and $F^{\beta}_{i}(\cdot)$ denote the $i$-th fully-connected layers for generating parameters of scaling ($\alpha$) and shifting ($\beta$) operators, respectively. Note that the dimensions of $\alpha_i$ and $\beta_i$ are equal to the channel number of the  intermediate feature maps in the $i$-th modulation module.

\subsection{Modulation in Base Networks}
\label{sec:Modu_Base}
In this section, we introduce our modulation strategies adopted in the generator and discriminator, which are illustrated in Figure \ref{figure_framework}. The parameters of modulation operations are generated by the condition networks.

For the generator, we impose controls on different features from different scales as well as within one scale. As we mentioned above, there are three scales in the generator. 
In different scales, the feature maps have different spatial sizes from high resolution to low resolution. 
In particular, for two successive scales: $scale_{m}$ and $scale_{m+1}$, we have:
\begin{align}
x_{m}^{\prime} = F^{G}_{m+1}(x_{m}) + x_{m}, \nonumber
\end{align}
where $F^{G}_{m+1}(\cdot)$ denotes the transformation for $scale_{m+1}$ in the generator $G$, $x_{m}$ is the feature maps in $scale_m$ before entering the $scale_{m+1}$, and $x_{m}^{\prime}$ is the features maps obtained after the addition of the feature maps from $scale_{m+1}$.

Then, we formulate the modulated scale fusion (MSF) on features from  $scale_{m}$ and $scale_{m+1}$:
\begin{align}
x_{m}^{\prime} = w_i * F^{G}_{m+1}(x_{m}) + x_{m}, \nonumber
\end{align}
where $w_i$ is the tuning weight generated by the $i$-th fully connected layer of the condition network for generator. Besides, the dimension of $w_i$ is the same as the number of feature maps. This controlling strategy could dynamically change the weights for lower-resolution features. Intuitively, when the degradation information of the input image is extremely mild, the tuning weights will be changed close to zero, since there is no need to receive rich contextual information from low resolutions.

Within one scale, we formulate modulated residual block (MRB) to achieve control. To be more exact, we have:
\begin{align}
x^{\prime}_i = w_i * f(x_i) +  x_i, \nonumber
\end{align}
where $f(\cdot)$ denotes the transformation in the residual branch, $x_{i}$ and $x^{\prime}_i$ are the original and modulated feature maps, respectively. $ w_i$ is the corresponding tuning weight.

In the discriminator, we adopt global feature modulation (GFM) to modulate the intermediate feature maps. To be more specific, we scale and then shift the intermediate features based on the condition. Practically, we perform modulation after batch normalization. In the $i$-th modulation module, we obtain the modulated feature maps by:
\begin{align}
x^{\prime}_i = \alpha_i * x_i + \beta_{i}, \nonumber
\end{align}
where $x_i$ and $x^{\prime}_i$ are the original and modulated feature maps. The multiplier $\alpha_{i}$ and addition operator $\beta_{i}$ are respectively generated by two independent fully connected layers from the condition network for discriminator.

\begin{table*}[]
	\vspace{-2.5em}
	\resizebox{\textwidth}{!}{
		\renewcommand{\arraystretch}{1.05}
		\setlength{\tabcolsep}{2pt}
\begin{tabular}{|c|c|lccccccccccccl|}
\hline
\multicolumn{2}{|r|}{Degradation Settings} &
  \multicolumn{6}{c|}{Single Degradation} &
  \multicolumn{8}{c|}{Two Degradations} \\ \hline
\multicolumn{2}{|r|}{blur} &
   &
  0 &
  0 &
  2 &
  4 &
  \multicolumn{1}{l|}{} &
  \multicolumn{1}{l}{} &
  1 &
  1 &
  2 &
  2 &
  4 &
  4 &
   \\
\multicolumn{2}{|r|}{noise} &
   &
  30 &
  50 &
  0 &
  0 &
  \multicolumn{1}{l|}{} &
  \multicolumn{1}{l}{} &
  15 &
  30 &
  30 &
  50 &
  30 &
  50 &
   \\
\multicolumn{2}{|r|}{condition vector} &
   &
  \multicolumn{1}{l}{[0.0, 0.6]} &
  \multicolumn{1}{l}{[0.0, 1.0]} &
  \multicolumn{1}{l}{[0.5, 0.0]} &
  \multicolumn{1}{l}{[1.0, 0.0]} &
  \multicolumn{1}{l|}{} &
  \multicolumn{1}{l}{} &
  \multicolumn{1}{l}{[0.15, 0.30]} &
  \multicolumn{1}{l}{[0.15, 0.60]} &
  \multicolumn{1}{l}{[0.50, 0.60]} &
  \multicolumn{1}{l}{[0.50, 1.00]} &
  \multicolumn{1}{l}{[1.00, 0.60]} &
  \multicolumn{1}{l}{[1.00, 1.00]} &
   \\ \hline \hline
Metric &
  Model Name &
  \multicolumn{14}{c|}{} \\ \hline
 &
  UGAN &
   &
  0.0490 &
  0.0957 &
  0.0484 &
  0.1331 &
   &
   &
  0.0697 &
  0.1140 &
  0.2068 &
  0.2557 &
  0.3154 &
  0.3510 &
   \\
\textit{\textbf{LPIPS $\downarrow$}} &
  CUGAN &
   &
  0.0522 &
  0.0966 &
  0.0525 &
  0.1463 &
   &
   &
  0.0714 &
  0.1178 &
  0.2019 &
  0.2498 &
  0.3075 &
  0.3350 &
   \\ \cdashline{4-7} \cdashline{10-15}
 &
  \textbf{} &
  \textbf{distance} $\downarrow$ &
  \textbf{0.0032} &
  \textbf{0.0009} &
  \textbf{0.0041} &
  \textbf{0.0132} &
  &
  &
  \textbf{0.0017} &
  \textbf{0.0038} &
  \textbf{-0.0049} &
  \textbf{-0.0059} &
  \textbf{-0.0079} &
  \textbf{-0.0160} &
   \\ \hline
 &
  UGAN &
   &
  0.0603 &
  0.0901 &
  0.0499 &
  0.1013 &
   &
   &
  0.0716 &
  0.0987 &
  0.1430 &
  0.1705 &
  0.1994 &
  0.2110 &
   \\
\textit{\textbf{DISTS $\downarrow$}} &
  CUGAN &
   &
  0.0639 &
  0.0964 &
  0.0525 &
  0.1081 &
   &
   &
  0.0727 &
  0.1003 &
  0.1392 &
  0.1657 &
  0.1982 &
  0.2162 &
   \\ \cdashline{4-7} \cdashline{10-15}
 &
  \textbf{} &
  \textbf{distance} $\downarrow$ &
  \textbf{0.0036} &
  \textbf{0.0063} &
  \textbf{0.0026} &
  \textbf{0.0068} &
  &
  &
  \textbf{0.0011} &
  \textbf{0.0016} &
  \textbf{-0.0038} &
  \textbf{-0.0048} &
  \textbf{-0.0012} &
  \textbf{0.0052} &
   \\ \hline
\end{tabular}
	}
	\setlength{\abovecaptionskip}{2.0cm}
	\caption{Quantitative LPIPS/DISTS results for GAN-based modulation testing on LIVE1. $\downarrow$ means the lower the better. LPIPS/DISTS distance are bolded to stress the modulation results.
	}
	\label{table1}
\end{table*}

\begin{table}[]\scriptsize
\setlength{\tabcolsep}{1.2mm}{
\begin{tabular}{lcccccccc}
\hline
\multicolumn{1}{r}{}      & \multicolumn{4}{c}{Single Degradation} & \multicolumn{4}{c}{Two Degradations} \\ \cline{2-9} 
\multicolumn{1}{r}{blur}  & 0       & 0      & 2      & 4      & 1      & 1      & 2      & 4      \\
\multicolumn{1}{r}{noise} & 30      & 50     & 0      & 0      & 15     & 30     & 50     & 50     \\ \hline
Unet (1.364M)             & 30.62   & 28.28  & 30.24  & 26.85  & 29.12  & 27.41  & 24.57  & 23.03 \\
CUnet (1.370M)            & 30.50   & 28.16  & 20.07  & 28.63  & 29.04  & 27.35  & 24.55  & 23.02 \\
\textbf{PSNR Distance $\downarrow$} &
  \textbf{0.12} &
  \textbf{0.12} &
  \textbf{0.17} &
  \textbf{0.22} &
  \textbf{0.08} &
  \textbf{0.06} &
  \textbf{0.02} &
  \textbf{0.01} \\ \hline
CResMD (2.50M) &
  \multicolumn{1}{l}{30.43} &
  \multicolumn{1}{l}{28.06} &
  \multicolumn{1}{l}{30.09} &
  \multicolumn{1}{l}{26.53} &
  \multicolumn{1}{l}{29.00} &
  \multicolumn{1}{l}{27.28} &
  \multicolumn{1}{l}{24.48} &
  \multicolumn{1}{l}{22.95} \\ \hline
\end{tabular}
}
\caption{Quantitative PSNR results for MSE-based modulation testing on CBSD68. The total parameter of the model is presented in the bracket. $\downarrow$ means the lower the better. PSNR distance are bolded to stress the modulation results.
}
\label{table2}
\vspace{-2em}
\end{table}

\subsection{Interpolation between modulation models.}
\label{sec:DNImetod}
The proposed CUGAN can achieve modulation for photo-realistic image restoration. 
To meet different user flavors, we use the deep network interpolation strategy (DNI) \cite{DNI} to achieve smooth transition between GAN and MSE effects.
Specifically, we first train a PSNR-oriented modulation model with MSE loss, and obtain the networks $G_{MSE}$ and $C_{MSE}$. 
For GAN training, we directly finetune all parameters of the PSNR-oriented modulation model, and obtain the networks $G_{GAN}$ and $C_{GAN}$. 
Then, we interpolate all the corresponding parameters of these two modulation models to obtain an interpolated modulation model, whose parameters are:
\begin{align}\nonumber
\theta^{interp}_{G} = (1-\alpha)\theta^{GAN}_{G} + \alpha\theta^{MSE}_{G}, \\\nonumber
\theta^{interp}_{C} = (1-\alpha)\theta^{GAN}_{C} + \alpha\theta^{MSE}_{C}, 
\end{align}
where $\theta^{interp}_{G}$, $\theta^{MSE}_{G}$ and $\theta^{GAN}_{G}$ are the parameters of the base networks $G_{interp}$, $G_{MSE}$ and $G_{GAN}$, respectively, $\theta^{interp}_{C}$, $\theta^{PSNR}_{C}$ and $\theta^{GAN}_{C}$ are the parameters of the condition networks $C_{interp}$, $C_{PSNR}$ and $C_{GAN}$, respectively, and $\alpha\in[0,1]$ is the interpolation coefficient.

\begin{figure*}[t]  
\centering  
\includegraphics[width=1.0\linewidth]
{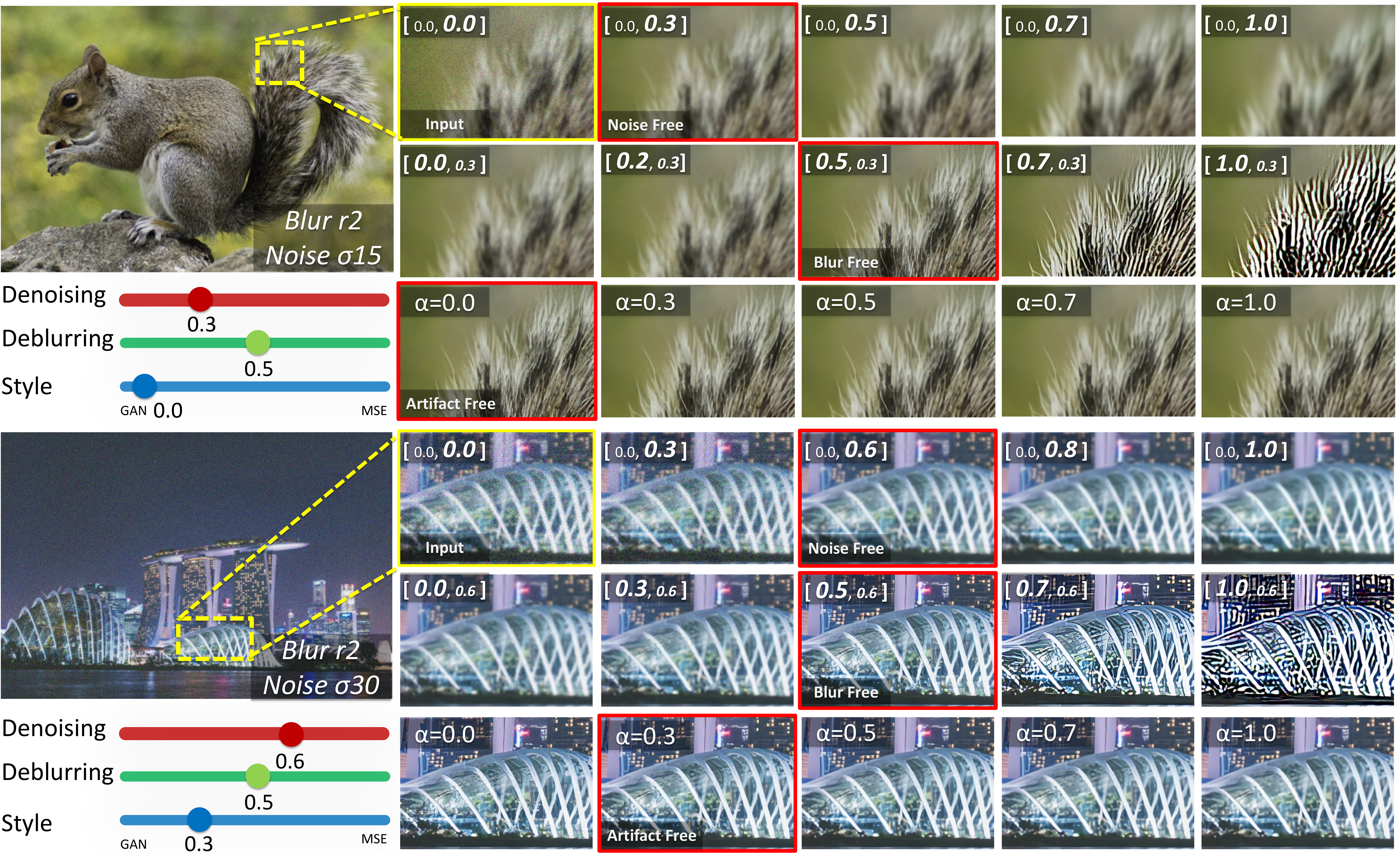}  
\caption{Qualitative results of Modulation for Image Restoration. For each group of images, three rows illustrate operations of denoising, deblurring, and interpolation, respectively. The colored-boxed images are the chosen results of the corresponding modulation process.}
\label{main_exp_figure}
\vspace{-1em}
\end{figure*}

For a given corrupted image $x$, we first specify an appropriate condition vector for photo-realistic image restoration. Then, we change the coefficient $\alpha$ from $0$ to $1$ to obtain smooth transition between GAN and MSE effects without artifacts (see Figure \ref{figure1}, \ref{main_exp_figure}).


\section{Experiments}


\subsection{Implementation Details}

We use DIV2K training dataset\cite{DIV2K}, which is a high-quality dataset and widely used in image restoration. We follow the work in \cite{cresmd} to generate the degraded input images.
Specifically, we add Gaussian blur and Gaussian noise sequentially to every training image in DIV2K dataset with random levels. 
The covariance range of Gaussian noise is $\sigma \in [0,50]$, and the range of kernel widths for $21 \times 21$ Gaussian blur is $ r \in [0,4]$. We uniformly sample the degradation levels with a stride of 0.1 and 1 for blur and noise, respectively. 
Besides, the degradation information for each degraded image is encoded into condition vector \textcolor{red}{$z$}.

The training process consists of two stages. 
First, we pre-train the PSNR-oriented modulation model by only using MSE loss, named as CUnet.
The base network $G$ and the corresponding condition network $C_{G}$ are jointly trained. 
The learning rate is initialized as $5\times 10^{-4}$ and decayed by a factor of 2 every $2\times 10^{5}$ iterations of update. 
The well-trained model CUnet will serve as the starting point for the next GAN training.
In the second stage, the base networks, $G$ and $D$, as well as the condition networks, $C_{G}$ and $C_{D}$ are jointly trained with the objective: 
\begin{align}
\mathcal{L}_{total} = \mathcal{L}_{percep} + 0.005\mathcal{L}_{GAN}+ 0.01\mathcal{L}_{MSE} \nonumber
\end{align}
where $\mathcal{L}_{MSE}$ is the MSE loss, $\mathcal{L}_{GAN}$ denotes the standard GAN loss, $\mathcal{L}_{percep}$ represents the perceptual loss. 
The perceptual loss is the L1 distance between two activated features obtained from VGG19-54$\footnote{We use pre-trained 19-layer VGG network, where 54 indicates features output by the 4th convolution before the 5th maxpooling layer}$.
The initial learning rate is set to $5\times 10^{-4}$ and will be halved at $[50k,100k,200k,300,400k]$. 
The number of mini-batch is set to 16, and the crop size is 64. For optimization, we use Adam \cite{adam} with $\beta_1$ = 0.9 and $\beta_2$ = 0.999. 
Moreover, we augment the training data with random horizontal flips and 90-degree rotations. 
All the models are implemented with the PyTorch framework and trained on NVIDIA 1080Ti GPUs.

\begin{figure*}[t]  
\vspace{-2.5em}
\centering  
\includegraphics[width=1.0\linewidth]
{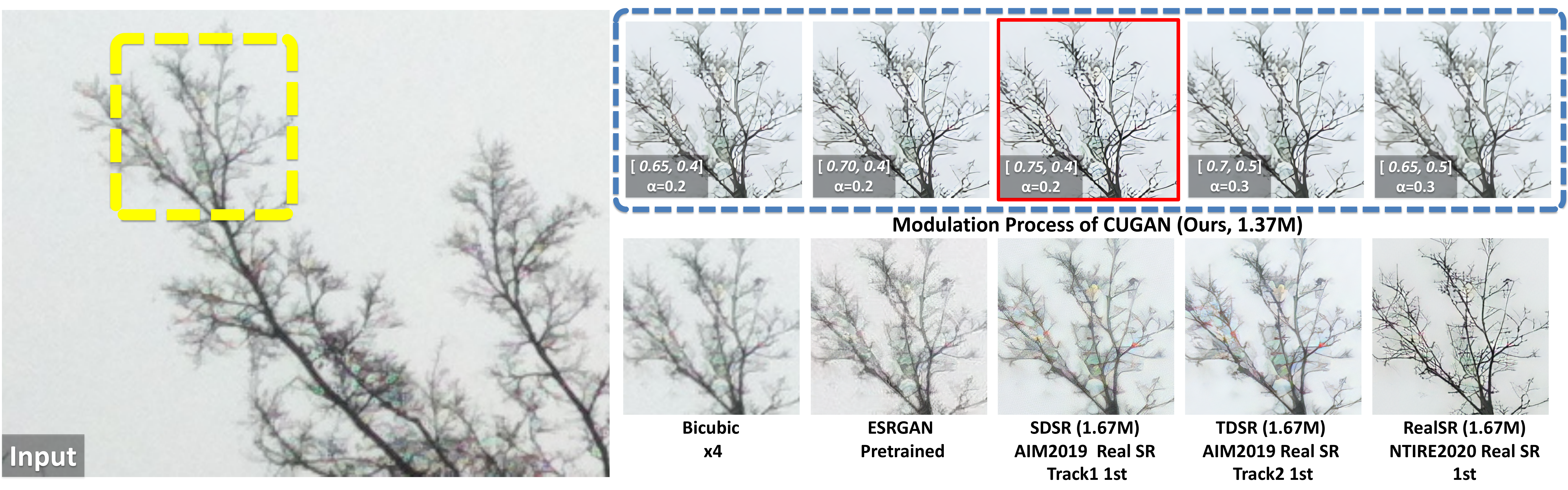}  
\caption{The qualitative comparison for blind SR task on DPED dataset \cite{DPED}. The images in the first row show the modulation process of CUGAN. The corresponding condition vectors and interpolation coefficients $\alpha$ are denoted.
The red-boxed image is the visually best result found by modulation. The number of total parameters of each model is presented in the bracket.}
\label{realworld_SR}
\vspace{-1em}
\end{figure*}

\subsection{Evaluation for Modulation Performance}
\textbf{Settings.}
LIVE1 \cite{LIVE1} and CBSD68 \cite{CBSD68} are chosen to evaluate our models.
We evaluate the modulation results on single degradation and two degradations. 
For single degradation, we choose denoising $\sigma30$, $\sigma50$ and deblurring $r2$, $r4$. For two degradations, we choose deblurring+denoising, [$r1$, $\sigma15$] [$r2$, $\sigma15$], [$r2$, $\sigma30$], [$r2$, $\sigma50$], [$r4$, $\sigma30$], [$r4$, $\sigma50$].
We train two baseline models, baseline-MSE named Unet (CUnet w/o condition network) and baseline-GAN named UGAN (CUGAN w/o condition networks), for each of chosen degradations.
Specifically, we remove the condition networks and only use the base networks to train these baseline models. 
The implementation details remain the same with CUnet and CUGAN, except that every baseline model is only trained on one specific degradation level.

Based on the evaluation results in PIPAL dataset\cite{pipal}, LPIPS \cite{lpips} and DISTS \cite{dists} have higher consistency with human ratings than NIQE, PI, and MA on images from the GAN model.
Therefore, in our experiments, PSNR is used to evaluate the performance of PSNR-oriented models, while DISTS and LPIPS are used to evaluate the GAN-based models. 
Given the ground truth images, we calculate the PSNR of CUnet and Unet and compute the PSNR distance.
Then we calculate LPIPS and DISTS of CUGAN and UGAN and compute the LPIPS distance as well as the DISTS distance. 
Lower PSNR/LPIPS/DISTS distance indicates better modulation results.

\textbf{Modulation for Image Restoration.}
\label{sec:Modu_Result}
Here, we evaluate the modulation performance of the proposed CUGAN quantitatively and qualitatively. 
The ideal performance of CUGAN and CUnet is to approach the performance of baseline models on every degradation level as close as possible.
In Table \ref{table1}, we evaluate the GAN-based modulation across various degradations on LIVE1 dataset.
For the single degradation, the LPIPS/DISTS distances are all smaller than 0.0041/0.0068. 
This result demonstrates that our CUGAN could achieve modulation with high quality. 
As for two degradations (blur+noise), the modulation performance of CUGAN is even better than UGAN on several degradations: [$r2$, $\sigma30$], [$r2$, $\sigma50$], [$r4$, $\sigma30$], [$r4$, $\sigma50$]. 
The quantitative results indicate that our CUGAN has better potential and ability on image restoration when degradations are more complex and severe.
The first two rows of Figure \ref{main_exp_figure} show the qualitative results, which present the smooth and stable transitions by modulation through the condition vector.
In Table \ref{table2}, we provide the PSNR-oriented modulation evaluated on the CBSD68 dataset. 
The PSNR distances for all degradations are below 0.22dB.
Furthermore, CUnet outperforms CResMD on almost all degradations. 
Specifically, CUnet surpasses CResMD on deblurring $r4$ with 0.32dB, and denoising $\sigma50$ with 0.22dB, which indicates a significant improvement. 
Moreover, CResMD contains 2.5 million parameters while ours contains 1.37 million parameters.
These results prove the effectiveness and superiority of our proposed multi-dimension modulation framework.

\begin{table}[]\scriptsize
\setlength{\tabcolsep}{4.8mm}{
\begin{tabular}{|c|c|c|c|c|}
\hline 
methods & a = 15 & a = 25 & a = 30 & a = 50 \\ \hline \hline
BM3D    & 31.08  & 28.57  & 27.76  & 25.62  \\ \hline
TNRD    & 31.42  & 28.92  & 27.66  & 25.97  \\ \hline
DnCNN-B & 31.61  & 29.16  & 28.36  & 26.23  \\ \hline
IRCNN   & 31.63  & 29.15  & 28.26  & 26.19  \\ \hline
FFDNet  & 31.63  & 29.19  & 28.26  & 26.29  \\ \hline
CUnet   & \textbf{34.53}  & \textbf{31.80}  & \textbf{30.89}  & \textbf{28.48}  \\ \hline
\end{tabular}
}
\caption{Comparison with different image denoising methods in term of PSNR on gray CBSD68. The highest PSNR on various noise degrees are bolded. 
}
\label{denoise}
\vspace{-2em}
\end{table}

\textbf{Smooth transition between MSE and GAN effects.}
The proposed CUGAN provides the default modulation results with GAN effects. Meanwhile, CUnet provides reconstruction style--MSE effect, which is consistent with high reconstruction accuracy (high PSNR). 
We apply linear interpolation for all parameters of CUnet and CUGAN to achieve transition between GAN and MSE effect. 
In Figure~\ref{main_exp_figure}, the third row illustrates the trade-off between these two effects. 
After we have obtained a photo-realistic restored image by modulation through the condition vector, we could further smooth the image by changing an interpolation coefficient $\alpha$ from $0$ to $1$. 
In the transition, users could make the unpleasing textures/artifacts gradually disappear to obtain an artifact-free output.

\subsection{Comparison on Image Denoising.}
To show the competitive denoising ability of the proposed model, we compare CUnet trained for Section \ref{sec:Modu_Result} with several state-of-the-art denoising methods: BM3D\cite{cbm3d}, TNRD\cite{TNRD}, DnCNN\cite{dncnn}, IRCNN\cite{IRCNN} , FFDNet\cite{ffdnet}.
We test them on gray CBSD68 using the mean PSNR as the quantitative metric.
To obtain degraded images, gaussian noise of different levels (e.g., 15, 25, 30, 40, and 50) are added to clean images.
Then, we modulate the condition vector corresponding to the noise level to obtain restored images.
Table \ref{denoise} provides the quantitative comparison, which presents the competitive performance of our method on image denoising.
CUnet provides competitive PSNR performance comparing with those state-of-the-art denoising methods.


\begin{figure}[t]
	\begin{center}
		\includegraphics[width=1.0\linewidth]{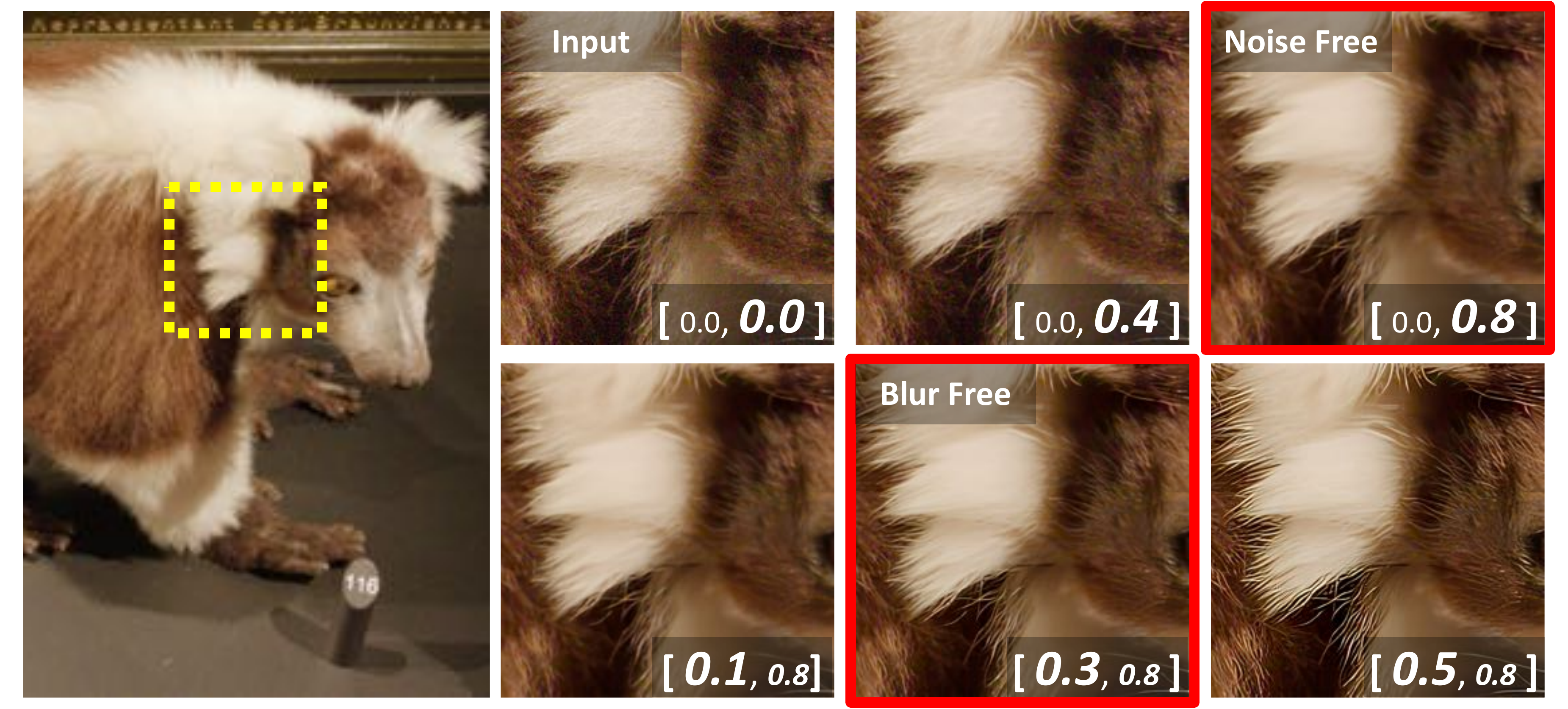}
	\end{center}
	\caption{Modulation for image restoration in real-world images. The input image with unknown noise is from NIND \cite{NIND} data set. Two rows present denoising and debluring, respectively.}
	\label{fig:real}
	\vspace{-1em}
\end{figure}

\subsection{Modulation on Real-World Images}
To further validate the generalization ability of the proposed CUGAN, we evaluate the effectiveness of modulation on real-world images SR and denoising. 

\textbf{Qualitative Results on DPED.}
In real-world image SR, the LR images usually have unknown degradations, such as complicated downsampling kernel and noise. 
Therefore, the non-blind GAN-based SR methods fail to produce satisfactory results. The proposed CUGAN could deal with this situation through modulation. 
In Figure \ref{realworld_SR}, given an LR input from DPED \cite{DPED}, we first up-sample it by bicubic interpolation. 
Then, we obtain the restored result by changing the condition vector and interpolation coefficient.
CUGAN has a competitive performance compared with those winner methods \cite{fritsche2019frequency,ji2020real} in the AIM2019 \cite{aim_realsr} and NTIRE2020 \cite{ntire_realsr} real-world image SR challenge with fewer parameters.

\textbf{Qualitative Results on NIND.}
As for real-world image denoising, visual results are provided in Figure \ref{fig:real}. 
The input image from a real-world denoising dataset NIND \cite{NIND} contains noticeable but unknown noise. 
We first use our CUGAN to eliminate the noise by changing the condition vector from $[0.0, 0.0]$ to $[0.0, 0.8]$. 
The restored image is noise-free but slightly over-smoothing. 
Then, we sharpen this restored image by changing the first element from $0.0$ to $0.3$. 
After the modulation process, we can obtain a noise-free as well as a non-blurry restored result.

\subsection{Ablation Study}
\label{section:ablation}
\begin{figure}[htbp]  
\vspace{-1.0em}
\centering  
\includegraphics[width=8.4cm]
{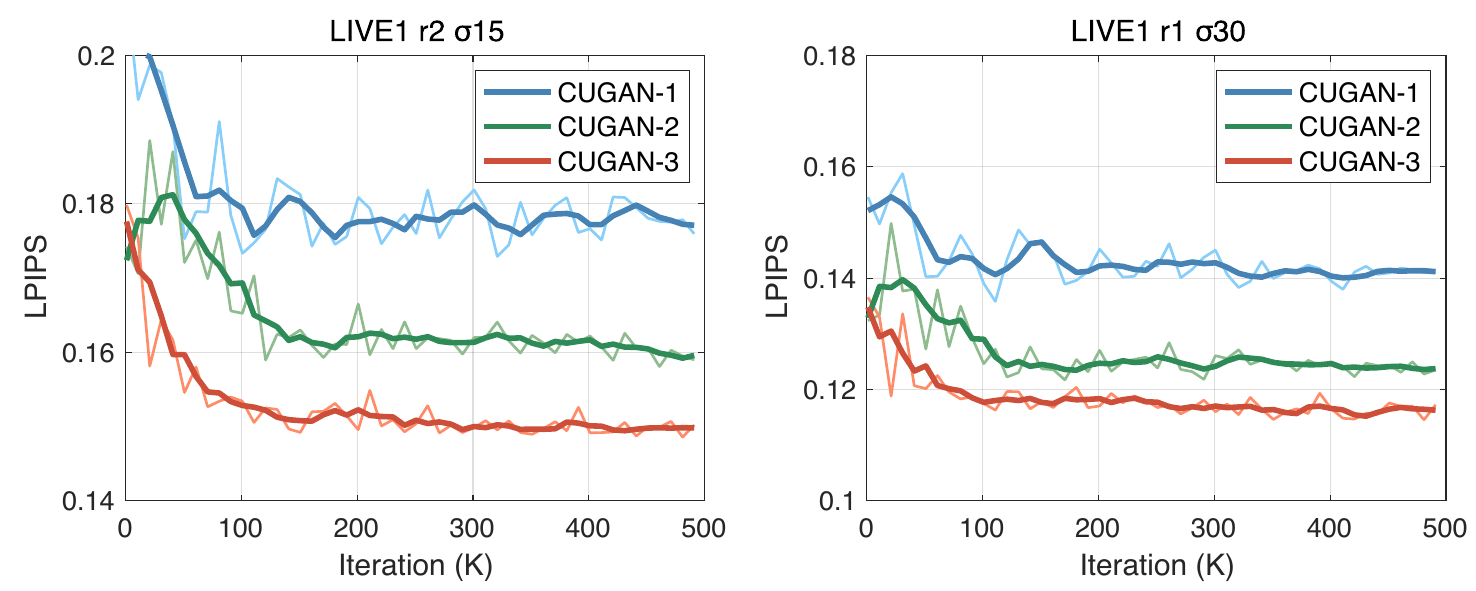}  
\caption{Convergence curves achieved by different models with different number of scale in term of LPIPS on LIVE1 dataset.} 
\label{fig:ablation1}
\end{figure}

\textbf{Effect of the multi-scale architecture in the generator.}
To verify the effectiveness of the proposed multi-scale architecture, we train three models CUGAN-1 (Plain Struture), CUGAN-2, and CUGAN-3 (Ours) under the same experimental settings.
Specifically, these three models have one scale, two scales, and three scales, respectively.  
Each model has 12 residual blocks, and each scale has the same number of residual blocks.
The convergence curves on LIVE1 dataset for degradations $[r1, \sigma30]$ and $[r2, \sigma15]$ are presented in Figure~\ref{fig:ablation1}.
We use LPIPS to evaluate the performance.
In Figure~\ref{fig:ablation1}, we can observe that CUGAN-3 (Ours) surpasses CUGAN-1 and CUGAN-2 by a large margin on those degradations.
Therefore, the multi-scale architecture is more effective in GAN-based modulation compared with plain networks. More convergence curves are in the supplementary.

\begin{figure}[htbp]  
\centering  
\includegraphics[width=8.8cm]
{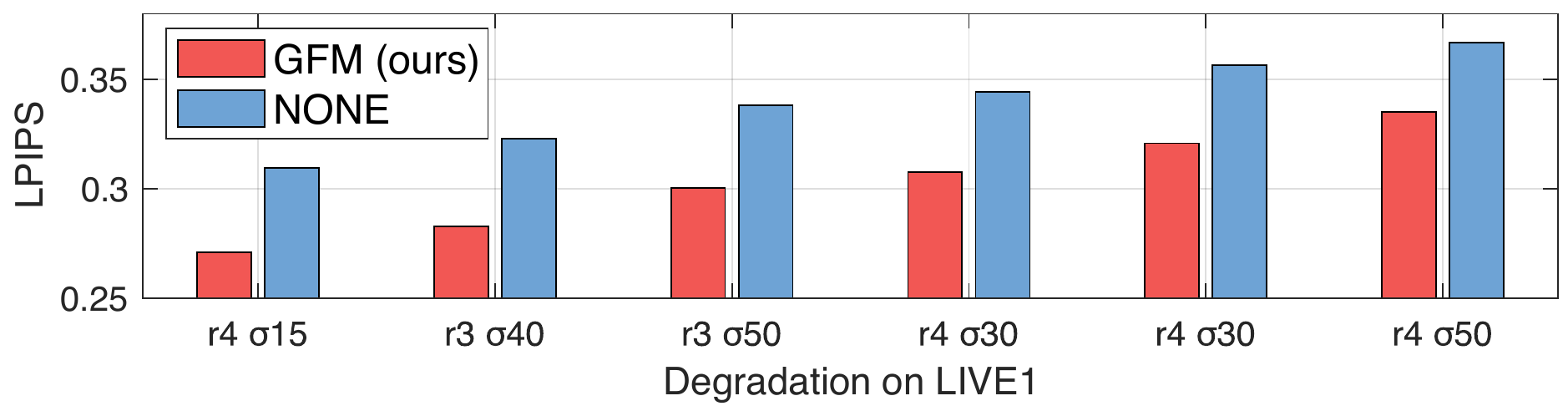}  
\caption{Results achived by CUGAN and CUGAN without GFM.} 
\label{fig:ablation2}
\vspace{-1em}
\end{figure}

\textbf{Effect of modulation in the discriminator.} 
In this section, we investigate the effectiveness of the modulation in the discriminator. 
Specifically, we train the GAN-based modulation model without the global feature modulation (GFM) in the discriminator and compare it with our CUGAN on different degradations evaluated by LIVE1 dataset. The performance of these two models in terms of LPIPS is shown in Figure \ref{fig:ablation2}. 
We can observe that our CUGAN equipped with GFM outperforms the other one on all those degradations. 
It is consistent with our assumption that the condition information is essential for the discriminator to make proper discrimination on restored images from different degradations.


\section{Conclusion}
In this work, we propose a novel GAN-based modulation framework named CUGAN.
CUGAN consists of a generator and a discriminator, which are all controlled by conditions vector.
The key idea of the CUGAN is introducing a modulation for users to freely control the strength of restoration and texture reconstruction with a photo-realistic effect.
We apply modulated scale fusion (MSF) and modulated residual block (MRB) in the generator to achieve interactive modulation.
Moreover, the Global Feature Modulation (GFM) is brought into the discriminator to ease the vanishing generator gradient.
Although CUGAN could realize modulation across multiple degradations, the modulation strategy can be more effective and efficient. 
Better solutions are expected for future research.

\textbf{Acknowledgements.}
This work was supported in part by the Shanghai Committee of Science and Technology, China (Grant No. 20DZ1100800), in part by the Na- tional Natural Science Foundation of China under Grant (61906184), Science and Technology Service Network Ini- tiative of Chinese Academy of Sciences (KFJSTSQYZX- 092), Shenzhen Institute of Artificial Intelligence and Robotics for Society.


{\small
\bibliographystyle{ieee_fullname}
\bibliography{egbib}
}

\end{document}